\pdfoutput=1
\documentclass[11pt]{article}

\usepackage[final]{acl}

\usepackage{times}
\usepackage{latexsym}

\usepackage[T1]{fontenc}
\usepackage[utf8]{inputenc}

\usepackage{microtype}

\usepackage{inconsolata}

\usepackage{xcolor}
\usepackage{array,multirow,graphicx}
\usepackage{float}
\usepackage{lipsum}

\title{Naming, Describing, and Quantifying Visual Objects in Humans and LLMs}

\author{Alberto Testoni \\ ILLC, University of Amsterdam \\  \texttt{a.testoni@uva.nl}
        \And  Juell Sprott \\ University of Amsterdam \\ \texttt{juell.sprott@student.uva.nl} \AND
        Sandro Pezzelle \\ ILLC, University of Amsterdam \\ \texttt{s.pezzelle@uva.nl}}
\begin{document}
\maketitle
\begin{abstract}
%Current Vision \& Language Large Language Models (VLLMs) achieve state-of-the-art performance on various tasks and are particularly effective in describing images and answering questions about them. 
While human speakers use a variety of different expressions when describing the same object in an image,
%(e.g., `a brown dog', `a cute dog on a sofa', etc.)
giving rise to a distribution of plausible labels driven by pragmatic constraints, the extent to which current Vision \& Language Large Language Models (VLLMs) can mimic this crucial feature of language use is an open question. This applies to common, everyday objects, but it is particularly interesting for uncommon or novel objects for which a category label may be lacking or fuzzy.
Furthermore, similar patterns of variation are observed among human speakers for highly context-sensitive expressions, such as the quantifiers `few' or `most'.
%Furthermore, humans show clear production preferences for highly context-sensitive expressions, such as the quantifiers `few' or `most'.
%Furthermore, human variability is commonly observed for highly context-sensitive expressions, such as the quantifiers `few' or `most'. 
In our work, we evaluate VLLMs (FROMAGe, BLIP-2, LLaVA) on three categories (nouns, attributes, and quantifiers) where humans show great subjective variability concerning the distribution over plausible labels, using datasets and resources mostly under-explored in previous work. %: naming common objects, naming uncommon objects, and assigning non-numerical quantifiers. 
%By sampling multiple times from the model, we compare the generated samples with human production variability. 
Our results reveal mixed evidence on the ability of VLLMs to capture human naming preferences at generation time: while some models are good at mimicking human distributions for nouns and attributes, all of them fail to assign quantifiers, a task that requires more accurate, high-level reasoning.

%, with some models being good at mimicking human distributions for nouns and attributes but all of them fail to assign quantifiers, a task that requires more high-level reasoning accurately.

%with all models failing in tasks that require high-level reasoning such as assigning quantifiers. 

%considerable differences across tasks and models. 

%Even when the models are not accurate, analyzing their variability allows us to gain valuable insights. 
%The datasets analyzed in this work represent a promising benchmark for further research in this direction. 
\end{abstract}

\section{Introduction}\label{sec:intro}

\begin{figure*}[t]
\centering
  \includegraphics[width=0.95\linewidth]{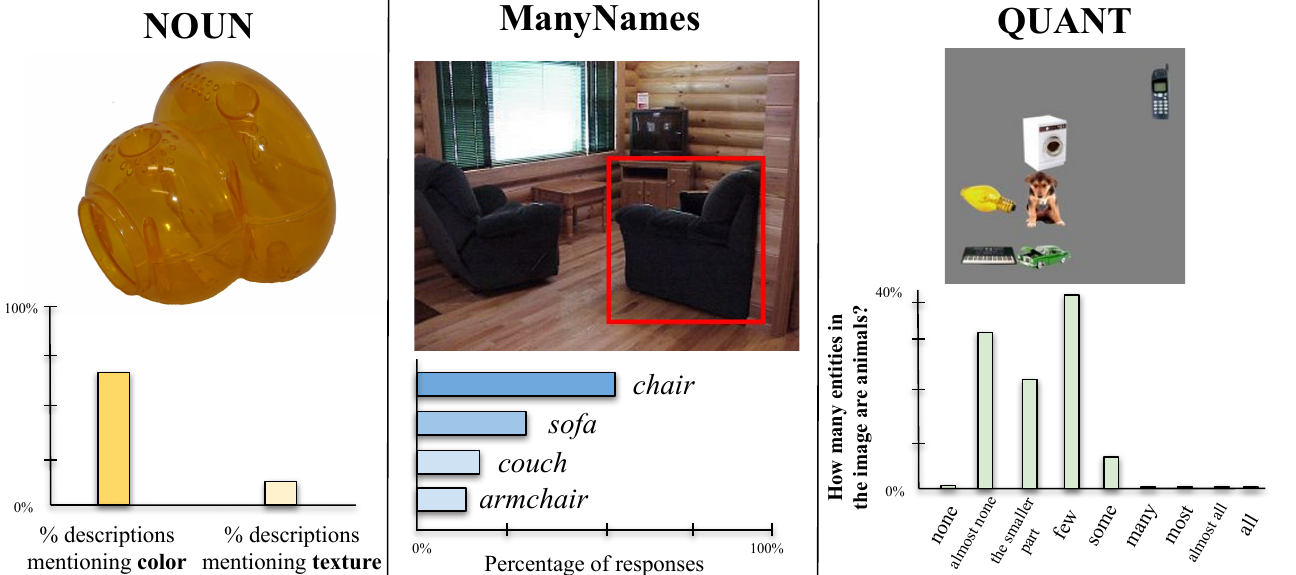}
  \caption{Datasets used in our experiments and distribution of human answers/labels. In NOUN (left), we focus on the frequency of color and texture attributes in the generated descriptions. In ManyNames (middle), each object is associated with the frequency of the nouns used to describe it. In QUANT (right), each image is associated with a probability distribution over a list of quantifiers that humans selected when answering the question `How many of the objects are animals?'. 
  }
  \label{fig:dataset_variability}
\end{figure*}

%\alberto{See the very end of the document for all the plots. }

Recent years have witnessed increasing popularity in the development of Large Language Models (LLMs) given their notable performance in following instructions, answering questions, and in many reasoning tasks, serving as general-purpose assistants \cite{huang-chang-2023-towards, zhao2023survey}. In parallel, a new generation of powerful Vision and Language LLMs (VLLMs) with excellent visual understanding and generation capabilities have emerged \cite{gan2022vision, li2023multimodal}. Rapidly, these models have outperformed previous approaches in many downstream tasks. In our work, we focus on the Natural Language Generation skills of powerful VLLMs by analyzing an important but under-explored problem, namely, their ability to capture human production variability (in terms of distribution over plausible labels/descriptions) in naming tasks.

Previous work highlighted that speakers display a wide range of variability when asked to utter sentences, resulting in inter-speaker variability but also variability over time for the same speaker \citep{levelt1993speaking, fan-etal-2018-hierarchical, alva2021suitability, takmaz2024describing}. In particular, in object naming, speakers may refer to objects appearing in a visual scene in many different ways \citep{graf2016animal}. Objects generally belong to multiple categories/super-categories, and all the lexicalized labels of such categories are valid \citep{brown1958shall}. However, although multiple labels are valid, humans pragmatically adapt their naming preferences depending on the context \citep{olson1970language, rohde2012communicating}, resulting in some labels being more frequently uttered than others. For instance, `mammal' is a correct label to describe a Gold Retriever, but pragmatically less likely than `dog'. Similarly, speakers tend to prefer sub-ordinate words like `car' instead of the potentially ambiguous super-ordinate word `vehicle' in case multiple vehicles appear in the image. In our work, we are interested in capturing both these two features: while many labels are equally valid and acceptable when naming or describing entities, these labels distribute according to a certain likelihood distribution.

%For instance, speakers tend to prefer sub-ordinate words like \textit{car, motorcycle} instead of the potentially ambiguous super-ordinate word \textit{vehicle} in case multiple vehicles appear in the image. 

%Being flexible with respect to the level of specificity represents a crucial pragmatic skill that allows humans to avoid ambiguities, select the most effective word in a given context, or simply display subjective preferences given the richness of human language. Despite the relevance of this phenomenon, it is not clear to what extent current V\&L LLMs can mimic human production variability. %This represents an essential step to assess the strengths and weaknesses of these models, and inspire future work. 

In our work, we investigate this issue, which has recently entered the NLP research community \citep{plank-2022-problem}, in three different production conditions.
%, where humans are shown to exhibit substantial variability. 
First of all, we consider the ManyNames dataset \citep{silberer-etal-2020-object, silberer-etal-2020-humans}, where annotators assign labels to describe common objects in images in a referential expression generation setting
%, which has a long tradition in multimodal NLP research 
\citep{yu2016modeling, kazemzadeh-etal-2014-referitgame}. We also explore two additional resources that have not received much attention within the NLP community and that allow us to broaden the horizons of this phenomenon. First, we analyze the NOUN dataset \citep{horst2016novel}, where speakers describe uncommon and novel objects: we focus on both the choice of the adjectives and how they distribute in the across-subject distribution. %Here, we focus on adjectives and attribute variability, given the inadequacy of category labels to describe these objects. 
Finally, we investigate human production variability arising from the context-sensitive nature of non-numerical quantifiers using the data collected by \citet{pezzelle2018probing}. %These adjectives (or adjectival phrases) describe object quantities and, given the same scene, multiple quantifiers represent a valid alternative. 
%Quantifiers have long been studied given their peculiar properties on the semantic and pragmatic level, and their anchoring in key skills that characterize human intelligence, such as the mechanisms of quantity estimation and comparison \citep{deschamps2015processing}.

%Inspired by \citet{giulianelli-etal-2023-comes}, who evaluated neural text generators against human production variability against uncertainty in text-only tasks, %(e.g., machine translation, storytelling, open-domain dialogue, etc.),
We evaluate three VLLMs (FROMAGe, BLIP-2, LLaVA) on the above-mentioned tasks in a zero-shot setting. We sample multiple times from the model using nucleus sampling, mimicking various human speakers, and compare the generated samples against human production patterns using different metrics (Jensen–Shannon divergence and Pearson's correlation, depending on the task at hand). Our results show that models weakly to moderately mimic human distributions in naming common and uncommon objects. Instead, all of them fail to mimic human distributions when selecting quantifiers, as highlighted by our in-depth analyses.

\section{Tasks and Datasets}\label{sec:tasks_datasets}

%We consider three tasks to investigate human production variability and compare it to models' variability. 

We use the images and corresponding human labels or descriptions from three datasets in English, that we briefly describe below.

\paragraph{NOUN} 
%Dataset by \citet{horst2016novel}. 
The Novel Object and Unusual Name (NOUN) dataset \citep{horst2016novel} contains 64 images of multipart, multicolored, and three-dimensional uncommon and novel objects. The dataset was originally created for behavioral studies on word learning and, to the best of our knowledge, it has not been used for NLP research.
%, representing unnameable stimuli. 
%The objects are multipart, multicolored, and three-dimensional objects. 
We focus on the \textit{naming task}, where participants were asked to answer the question ``What would you call this object?''. The answers are sentences like: `a plastic object with red stuff on top'. For each object, the proportion of colors (e.g., `red', `bronze') and textures (e.g., `soft', `rough') was calculated as the number of attributes given the number of responses. An example from the dataset is reported in Figure \ref{fig:dataset_variability} (left), together with the ratio of colors and textures in human responses. In NOUN, we examine human production preferences on a high level, by looking at the frequency according to which certain adjectives (related to color and texture attributes) are used.
%The authors also collected additional information, such as the perceived novelty and familiarity of the objects, but we do not use these pieces of information in our study. 
%To compute texture and color saliency, we create a list of words and perform string-matching with the generated answers (details in Appendix \ref{appendix:noun}).

\begin{figure*}[t]
\centering
  \includegraphics[width=1\linewidth]{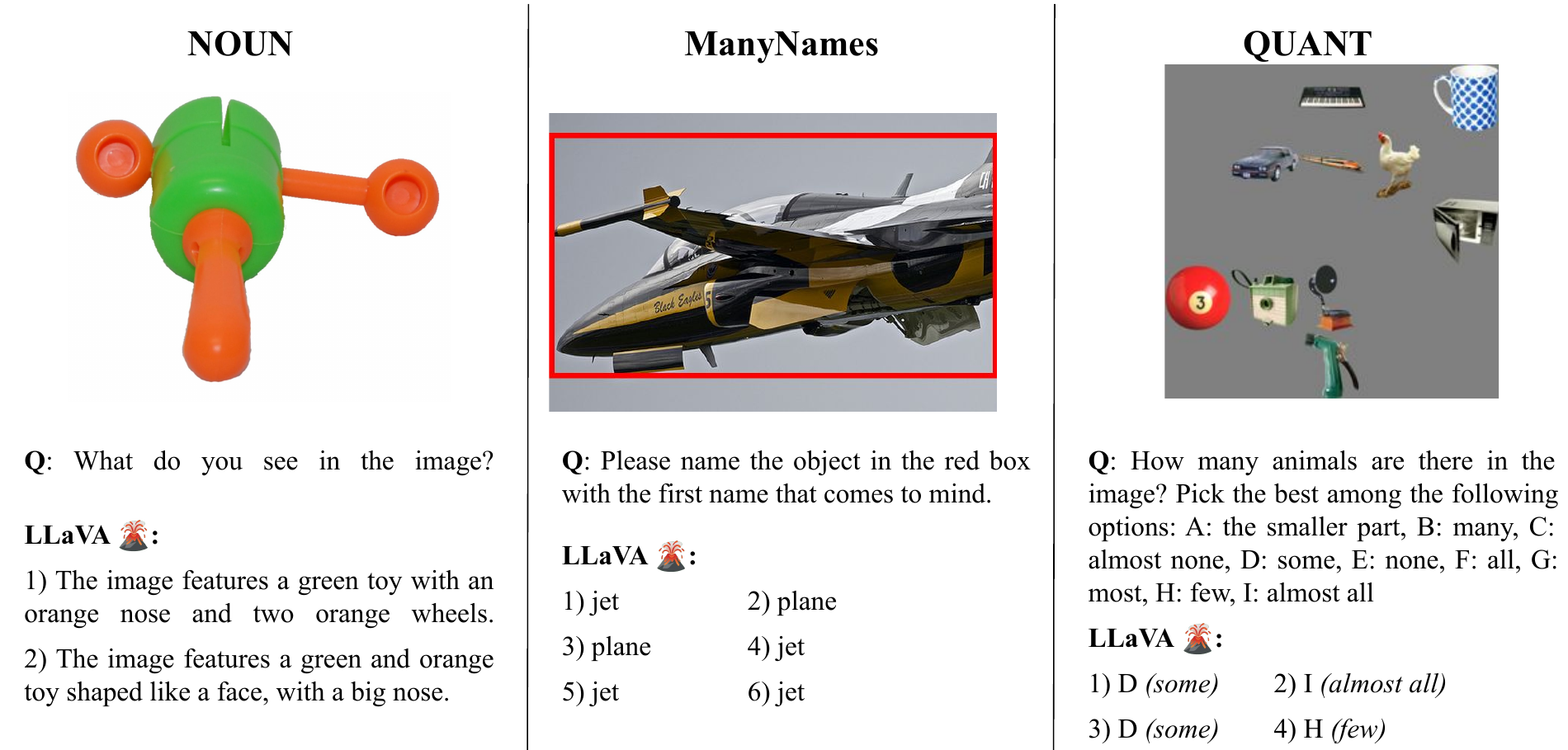}
  \caption{Examples of the output generated by LLaVA (multiple samples with nucleus sampling decoding) for the three tasks analyzed in our work. For each task, a sample of the answers provided by the model is displayed. For space constraints, we only report a few random samples for each task.}
  \label{fig:examples_models}
\end{figure*}

\paragraph{ManyNames} %Dataset by \citet{silberer-etal-2020-object, silberer-etal-2020-humans}. 
In ManyNames \citep{silberer-etal-2020-object, silberer-etal-2020-humans}, the authors collected names for 25K objects appearing in real-world images from VisualGenome \citep{krishna2017visual} by asking human annotators to generate a name for them. Each object (highlighted by a red box in the image) is associated with an average number of 35.3 annotations. More than 90\% of the objects are associated with more than one unique label (5.7 average name types per object). An example is shown in Figure \ref{fig:dataset_variability} (middle). When describing the object in the red box, most annotators referred to it as `chair', while around 30\% said `sofa', and the remaining ones used `couch' and `armchair'. The images in ManyNames are classified into 7 domains (e.g., vehicles, people, animals, etc.): for computational constraints, we evaluated 300 randomly sampled objects from each domain. Different from NOUN, we examine production preferences on a more fine-grained level using the actual distribution over multiple labels. %\alberto{Note: for computational constraints and limited resources, we took a subset of the images for ManyNames.} 

\paragraph{QUANT} To study how quantifiers are used when referring to quantities grounded in images, \citet{pezzelle2018probing} introduced a dataset of visual abstract scenes containing a variable number of animals and artifacts and asked human participants to answer the question ``How many of the objects are animals?". Participants could select the answer from a list of nine pre-selected quantifiers: `none', `almost none', `the smaller part', `few', `some', `many', `most', `almost all', and `all'. The authors used images with 17 different proportions of animals and artifacts (ranging from 0\% to 100\%). 
%For each proportion, the authors obtained a probability distribution over all quantifiers by aggregating participants' selection. For instance, considering the image in \ref{fig:dataset_variability} (right - 1 animal and 5 artifacts), most participants described the number of animals in the scene using the quantifiers \textit{almost none, the smaller part, few}.%, with \textit{few} being the most frequently selected. 
%A few participants selected the quantifier \textit{some} and none of them the remaining quantifiers. 
In our work, we tested 50 images for each of the 17 proportions in the dataset, resulting in a total number of 850 images.\footnote{The actual images used in our experiment come from \citet{testoni-etal-2019-quantifiers}, which built a large-scale dataset using the stimuli and pipeline by \citet{pezzelle2018probing}.}

\section{Experiments}\label{sec:experiments_results}

\subsection{Generation} In our work, we test the performance of three models in a zero-shot setting: BLIP-2 \citep{li2023blip}, FROMAGe \citep{koh2023grounding}, and LLaVA 1.5 \citep{liu2023visual, liu2023improved}. All three models can be prompted for zero-shot generation. Additional details are discussed in Appendix \ref{appendix:models}.
%\subsection{Prompting and Generation} 
For each of the three tasks described in Section \ref{sec:tasks_datasets}, we used prompts that resembled the instructions provided to human annotators during the dataset collection. ManyNames: \textit{Q: Please name the object in the red box with the first name that comes to mind. A:}. NOUN: \textit{Q: What do you see in the image? A:}. QUANT: \textit{Question: How many animals are there in the image? Pick the best among the following options: }, followed by the list of the nine quantifiers, each associated with a letter (from A to I). The ordering of the quantifiers is randomized at each inference step. Although investigating several variations of the above-mentioned prompts is beyond the scope of the paper, we discuss some insights on this aspect in Appendix \ref{appendix:generation}. 
We sample multiple times from each model using nucleus sampling decoding \citep{holtzman2019curious}, with $p=0.9$, $t=0.5$ (different hyperparameter configurations did not significantly affect the overall results, as discussed in Appendix \ref{appendix:generation}). For each task, we sample the model 20 times and filter out ill-formed answers, such as empty strings or question repetitions. After filtering, we randomly take 10 generations per image for ManyNames and NOUN, and 15 for QUANT. In this way, we have the same number of generations for each image/object. Some examples of the output generated by LLaVa for the three tasks analyzed are reported in Figure \ref{fig:examples_models}. We release our code at: \url{https://github.com/albertotestoni/ndq_visual_objects}.

\begin{figure}[t]
\centering
  \includegraphics[width=1\linewidth]{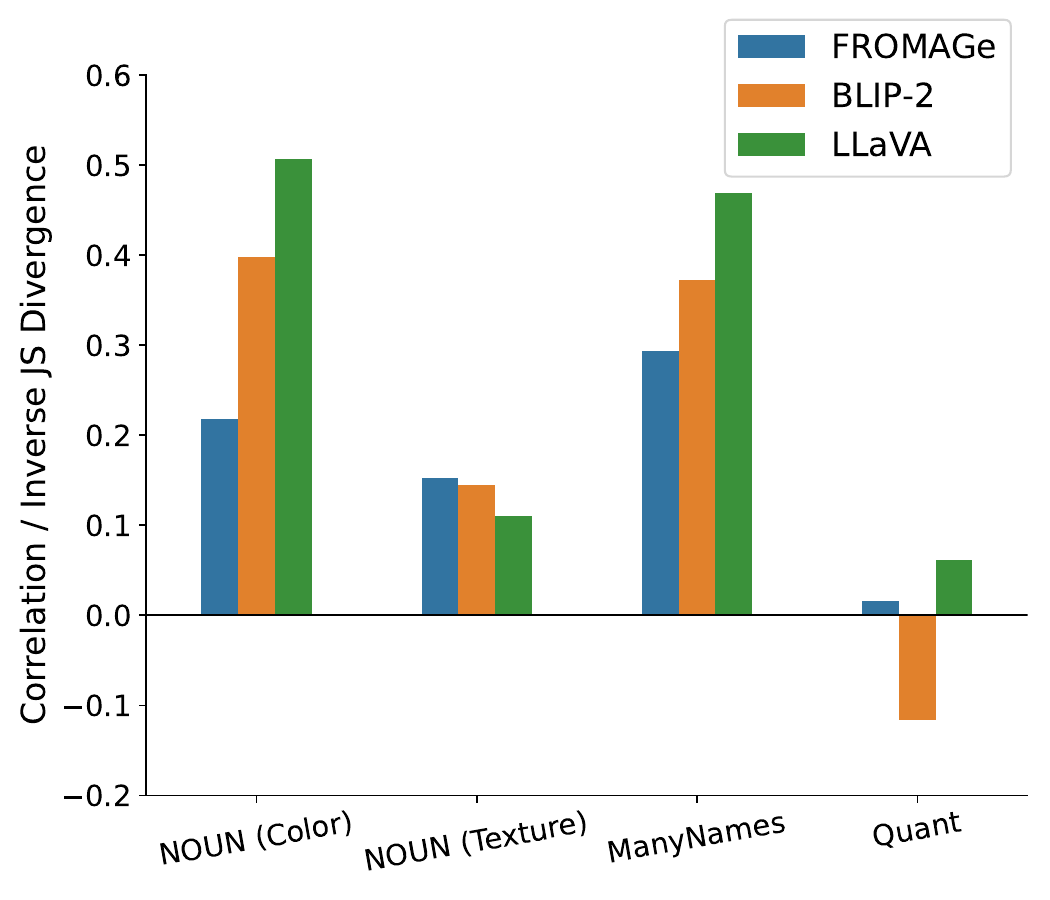}
  \caption{For NOUN and QUANT, the plot shows the correlation between human responses and model samples. %The correlation results are not statistically significant ($p<0.05$) for texture saliency in NOUN. 
  For ManyNames, it shows the inverse JS divergence between the frequency of the nouns chosen by annotators and the ones generated by the model.}
  \label{fig:results}
\end{figure}

\subsection{Evaluation}

Each object in \textbf{NOUN} is associated with color and texture saliency, i.e., how often speakers described the object using these attributes. We use a string-match approach (see Appendix \ref{appendix:noun}) to analyze the model output and compute color and texture saliency. We then compute the Pearson's \textit{r} correlation between human and model saliency, considering all objects.

Each object in \textbf{ManyNames} is associated with $H$ unique nouns assigned by human annotators and $M$ unique nouns sampled from the model output, together with their frequency. Given $A=H \cup M$, we construct two term-frequency vectors for human and model output, $h$ and $m$, respectively, with $|h|=|m|=|A|$. Each noun in $A$ is mapped to a unique position in $h$ and $m$ and each vector is filled with its normalized frequency. We evaluate the models by computing the inverse Jensen–Shannon (JS, bounded between 0 and 1) divergence \citep{lin1991divergence} between $h$ and $m$. See Figure \ref{fig:toy_example} in the Appendix for an example.
%\alberto{improve notation. Maybe add an example to the appendix?}

Each image in \textbf{QUANT} is associated with a probability distribution over 9 quantifiers, depending on the proportion of animals and artifacts. From the model outputs, we extract the relative frequency of each quantifier and compute Pearson's \textit{r} correlation with the human distribution. We then average the correlation results over all images. Correlation is bounded between -1 and 1. Higher is better for all the metrics.

\subsection{Results} 
As we can observe from Figure \ref{fig:results}, the results for ManyNames and NOUN (color saliency) show a clear trend: all the models correlate, to some extent, with human production, with LLAVA obtaining the highest correlations for both tasks (around 0.5) and significantly outperforming (t-test, $p<0.01$) both BLIP2 and FROMAGe.\footnote{Appendix \ref{appendix:mn} shows per-domain results for ManyNames.}
%the models moderately mimic human production variability, with LLaVA significantly outperforming (t-test, $p<0.01$) BLIP-2 and FROMAGe (Appendix \ref{appendix:mn} presents the per-domain results for ManyNames). 
These findings align with previous work showing the primacy of LLaVA over other models \citep{liu2023visual, liu2023improved}. However, the remaining tasks show critical weaknesses for all models. First, none of the models achieve a statistically significant correlation for texture saliency (all have $p>0.05$). We conjecture that texture attributes are less common for the models compared to colors, and thus they may be less accurate when generating them: we leave an in-depth analysis of this issue for future work. %, showing that they do not mimic human production variability in describing novel objects using texture attributes. 
Despite the correlation results being similar across models, our manual inspection reveals interesting differences: while the low performance of FROMAGe is due to an under-generation of texture attributes, the opposite is true for LLaVA, with BLIP-2 being more flexible in terms of texture attribute generation but not aligned with human variability (see Figure \ref{fig:noun_example} for an example). Finally, all models show almost no correlation in assigning quantifiers to visual scenes, highlighting a severe limitation of all models on this task.
%Finally, all models show almost no correlation or negative correlation in assigning quantifiers to visual scenes. This surprising result shed light on a severe weakness that characterizes all models in performing a task that mirrors an important feature of language use.
We scrutinize this issue in the following Section.
%In the following Section, we have a closer look at the model performance in this task. 

\begin{figure}[t]
\centering
  \includegraphics[width=1\linewidth]{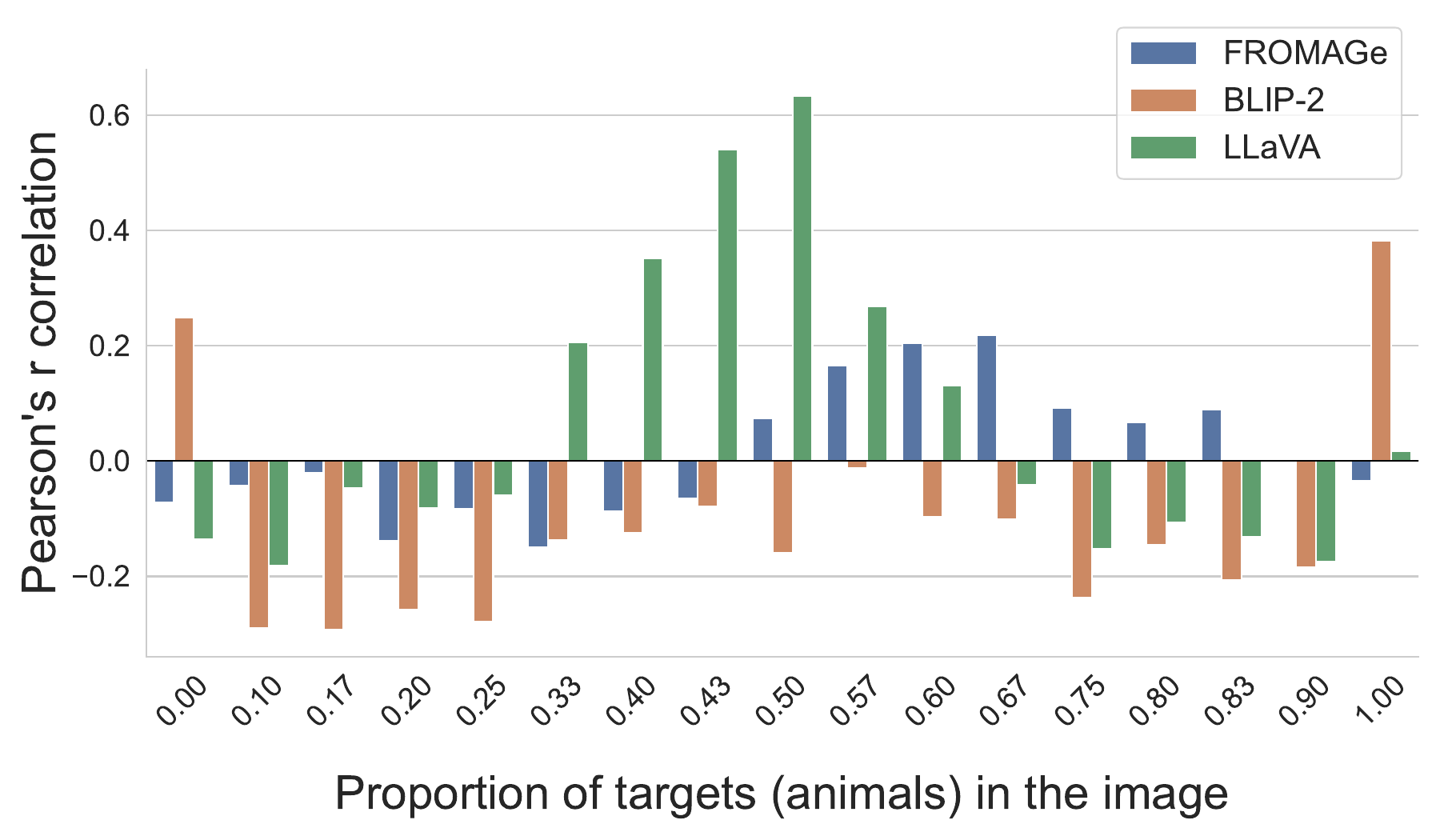}
  \caption{Pearson correlation results (y-axis) broken down by proportion of targets (animals) in the image (x-axis) in the QUANT task and dataset.}
  \label{fig:correlation_per_proportion}
\end{figure}

\section{\textit{The Curious Case} of Quantifiers}\label{sec:curious_case_quantifiers}

We run some analyses to investigate the poor performance of all models in the QUANT task. First of all, we acknowledge that the multiple-choice prompting used in QUANT is different and more complex than the prompts used for the other datasets. Still, it is unlikely that this is the main reason behind the poor performance of all models. 

%Even though all models show poor performance in the QUANT task (see Figure \ref{fig:results}), we believe that inspecting their variability allows us to gain valuable insights. 
Figure \ref{fig:correlation_per_proportion} shows the correlation results broken down by the proportion of animals in the image. We observe that even though the overall correlation results are similar across models (Figure \ref{fig:results}), they perform quite differently depending on the proportion of animals in the scene. While BLIP-2 performs relatively well on the `extreme' proportions (no animals or all animals in the image, when speakers generally choose the quantifiers `none' and `all', respectively), LLaVA excels at intermediate proportions, and FROMAGe performs better on proportions above 50\%. Can we conclude that models properly handle the task of assigning the most likely quantifiers for some proportions? These results evoke two hypotheses: (a) The models are capable of selecting plausible quantifiers only for some proportions or, vice versa, they understand only some of the quantifiers analyzed; (b) The models have a bias towards some specific quantifiers, regardless of the proportion of targets in the scene, leading to a decent perform on some proportions as a side effect. Our additional analyses, reported in Figures \ref{fig:density_plots} and \ref{fig:frequency_quantifiers} in the Appendix, support hypothesis (b): FROMAGe has a strong bias towards selecting the quantifier `many'; BLIP-2 frequently selects the \textit{extreme} quantifiers `none' and `all', and its selection is not influenced by the proportion of targets; LLaVA has a bias towards selecting the quantifier `some', regardless of the proportion of targets. 

%Overall, our results reveal that each model has a bias towards a different quantifier but all of them are not even close to successfully mastering this task. 
To further shed light on this result, we qualitatively assess the `counting' skills of the models, a crucial skill to succeed in assigning quantifiers. As the examples in Figure \ref{fig:counting} in the Appendix illustrate, all models struggle to successfully count how many animals appear in the image. We hypothesize that the reason for the poor performance in assigning quantifiers lies in the quantity estimation and comparison skills of the models. This observation is in line with recent research investigating the poor `counting' skills of current models \citep{paiss2023teaching}.
%We double-check that the models correctly process the visual input: when asked to generate a caption, they correctly output a list of the objects appearing in the image.

\section{Conclusion}\label{sec:conclusion}

While human speakers exhibit a wide range of human production variability in naming tasks, mirroring pragmatic constraints and subjective preferences, it is not clear to what extent VLLMs can mimic this peculiar trait of language use. 
%In our work, we take a first step in this direction.
%by evaluating three models (FROMAGe, BLIP-2, and LLaVA) on different tasks. 
%We use the ManyNames dataset to test models on naming common objects, and we propose to exploit additional under-explored resources (NOUN and the data collected on a study on quantifiers) to 
%We investigate different dimensions of production variability in three tasks: 
In our work, we investigate this issue in three tasks: naming common objects, naming novel objects, and assigning quantifiers. Our results reveal that best-performing models achieve a moderate correlation with human patterns in some tasks (object names and color terms). However, all models dramatically fail when assigning quantifiers, the only production setup that requires some form of reasoning, i.e., the ability to reason over sets of objects and process quantities. Based on our analyses, we hypothesize that the reason behind this failure stems from the poor ``counting'' skills of the models.

%a simple task that is rooted in core mechanisms of human intelligence.
% such as quantity estimation and comparison. 
%Still, we demonstrate that inspecting the output variability, even when the models are not accurate, allows us to gain valuable insights into biases and highlight the models' weaknesses. 
%Our work paves the way for further research in this direction, highlighting the suitability of the tasks analyzed. 

%Our analyses reveal that LLMs are biased towards predicting some specific quantifiers, regardless of the proportion of targets in the scene. Our results pave the way for further research in this direction, highlighting the relevance of the phenomena studies in the paper and the suitability of the tasks and datasets analyzed. \alberto{I think we should highlight more clearly the relevance of the results and come up with some hypotheses that could explain them.}

\section*{Limitations}\label{sec:limitations}
In the following, we discuss some limitations of our study that may inspire follow-up work in this direction. The poor performance on the quantification tasks may stem from the higher complexity of the prompt used (multiple choice prompting). Even though in our paper we discuss how analyzing the output variability allows us to gain valuable insights even when the model is not accurate, we can not rule out the possibility that a simpler prompt may lead to more accurate results. As an initial step, we used a prompt that corresponds to the instruction provided to the participants of the original experiment in \citet{pezzelle2018probing}. In Appendix \ref{appendix:generation} we discuss the effect of re-phrasing the original prompt instructions.

Moreover, it is worth noting that the human production variability analyzed in our experiments is obtained by aggregating data coming from multiple speakers. Even though we do aim at this, we acknowledge that it is unlikely that one single model can mimic such a rich variability. Our study is more focused on understanding \textit{to what extent} current Vision \& Language LLMs can mimic this feature, showing the suitability of some tasks and datasets not explored in previous work. 

Finally, we computed the color and saliency feature for NOUN using a string-matching approach based on a manually defined list of keywords (as described in Appendix \ref{appendix:noun}. We acknowledge that this approach may underestimate the color and texture saliency in the model output. Although in this case, the small size of the dataset allowed us to verify that this is not the case, we believe that it is important to take this point into account when running experiments on a larger scale. Moreover, as a limitation of the NOUN dataset (and not of our experimental setup), we do not have access to the actual color and texture labels used by human participants during the dataset collection. For this reason, in NOUN we do not consider the actual distribution of the attributes used by human speakers but just their overall frequency.

\section*{Acknowledgments}
We would like to thank the Dialogue Modelling Group (DMG) at the University of Amsterdam for their feedback and support at the different stages of this work. We thank Brent Brakenhoff for his input on some preliminary experiments with the ManyNames dataset. Alberto Testoni is supported by funding from the European Research Council (ERC) under the European Union’s Horizon 2020 research and innovation programme (grant agreement No.~819455, PI R. Fernández).

\bibliography{anthology,custom}

\appendix

\section{Appendix}
\label{sec:appendix}

\subsection{ManyNames Appendix}\label{appendix:mn}

Figure \ref{fig:frequency_quantifiers} shows the results on the ManyNames dataset broken down by the image domain. LLaVA outperforms other models in most of the domains, but for \textit{clothing} and \textit{people} it is comparable to BLIP-2. Note that all model reach have the poorest performance on these two domains. As highlighted by \citet{silberer-etal-2020-object, silberer-etal-2020-humans} and confirmed by our manual inspection, models confuse people and clothing objects much more frequently than humans do. ManyNames is licensed under Creative Commons Attribution 4.0 International.

\subsection{NOUN Appendix}\label{appendix:noun}

We define the following list of color and texture attributes to analyze the samples generated by the model with a string-matching approach.

\textbf{Colors} = [``Red'',    ``Orange'',    ``Yellow'',    ``Green'',    ``Blue'',    ``Purple'', ``Pink'',    ``Brown'',    ``Gray'',    ``Black'',    ``White'',    ``Beige'', ``Turquoise'',    ``Teal'',    ``Magenta'',    ``Lavender'', ``Indigo'',    ``Maroon'',    ``Gold'',    ``Silver'', ``Bronze'',    ``Copper'',    ``Olive'',    ``Navy'', ``Sky blue'', ``Cream'',    ``Peach'',    ``Rose'',    ``Fuchsia'',    ``Coral'',    ``Mint'', ``Chartreuse'',    ``Salmon'',    ``Sienna'',    ``Slate'',    ``Tan'', ``Crimson'',    ``Ivory'',    ``Khaki'',    ``Lilac'',    ``Mauve'', ``Mustard'',    ``Rust'',    ``Scarlet'',    ``Tangerine'', ``Vermilion'',    ``Violet'',    ``Wheat'',    ``Brick red'',    ``Caramel''] 

\textbf{Textures} = [``Smooth'',    ``Rough'',    ``Fuzzy'',    ``Soft'', ``Hard'', ``Bumpy'',    ``Slick'',    ``Sticky'',    ``Grainy'', ``Sandy'',    ``Slippery'',    ``Jagged'',    ``Sharp'', ``Coarse'',    ``Silky'',    ``Velvety'',    ``Wet'', ``Dry'',    ``Glossy'',    ``Matte'',    ``Sparkly'', ``Metallic'',    ``Wooden'',    ``Leathery'', ``Plastic'',    ``Rubber'',    ``Furry'', ``Woolly'',    ``Feathery'', ``Smooth'',    ``Satin'',    ``Lace'',    ``Crochet'',    ``Knitted'', ``Embroidered'',    ``Linen'',    ``Silk'',    ``Velvet'',    ``Suede'', ``Corduroy'',    ``Denim'',    ``Felt'',    ``Tweed'',    ``Mesh'', ``Hairy'',    ``Crisp'',    ``Crumbly'',    ``Flaky'',    ``Puffy'',    ``Spongy'', ``Crunchy'',    ``Chewy'',    ``Gummy'',    ``Slimy'',    ``Starchy'',    ``Syrupy'', ``Icy'',    ``Rocky'',    ``Stony'',    ``Sandy'',    ``Peppery'',    ``Salty'', ``Sour'',    ``Sweet'',    ``Tangy'',    ``Tart'',    ``Spicy'', ``Herbaceous'',    ``Earthy'',    ``Mossy'',    ``Woody'', ``Smoky'',    ``Smokey'',    ``Rusty'',    ``Corroded'',    ``Weathered'', ``Rugged'',    ``Smooth'',    ``Polished'',    ``Shiny'',    ``Gleaming'', ``Dull'',    ``Muddy'',    ``Cloudy'',    ``Milky'',    ``Transparent'', ``Translucent'',    ``Opaque'']

\begin{figure*}[]
\centering
  \includegraphics[width=1\linewidth]{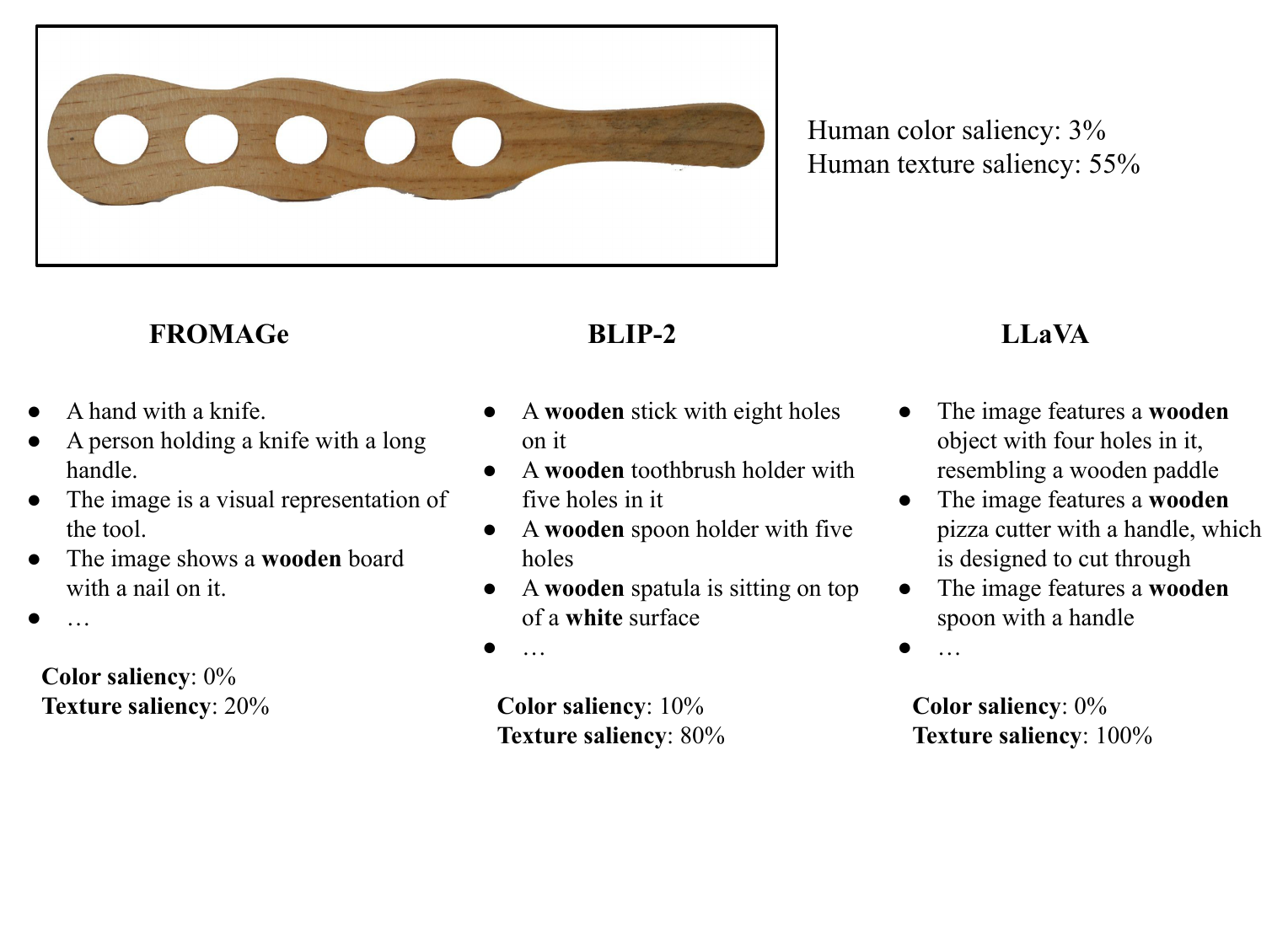}
  \caption{Example of the model output in the NOUN dataset.}
  \label{fig:noun_example}
\end{figure*}

Figure \ref{fig:noun_example} shows an example of the output of different models, together with their color and texture saliency as well as human saliency values. NOUN is released without a specific license. 

\begin{figure}[]
\centering
  \includegraphics[width=0.95\linewidth]{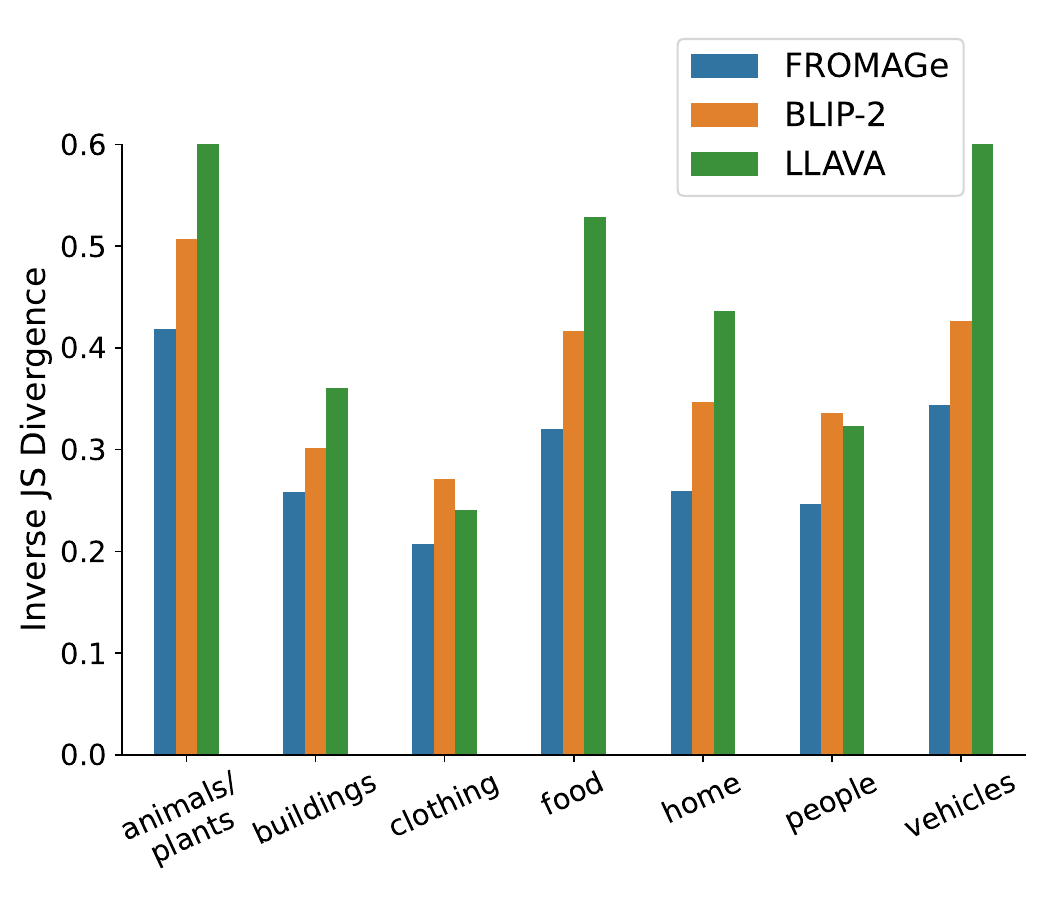}
  \caption{Inverse Jensen–Shannon divergence broken down by the image domain in ManyNames.}
  \label{fig:mn_distances_per_domain}
\end{figure}

\begin{figure*}[]
\centering
  \includegraphics[width=1\linewidth]{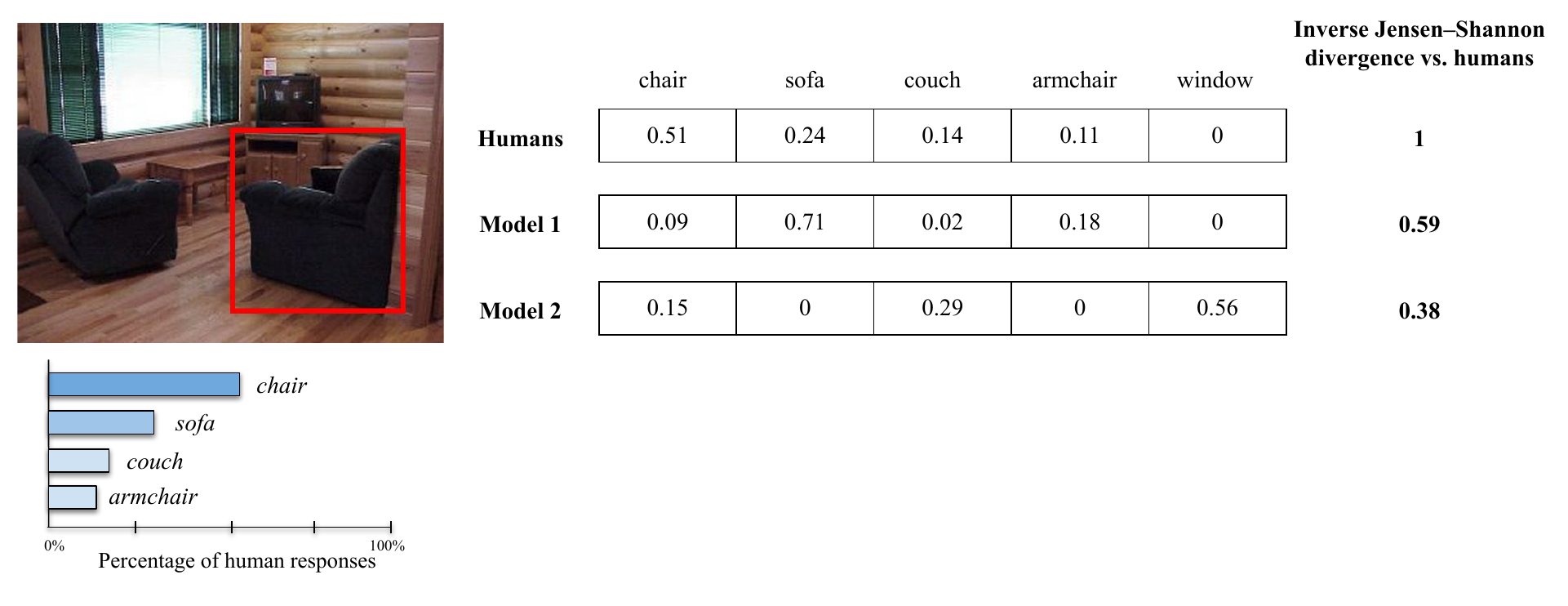}
  \caption{Toy example to show how the effect of over-generating a noun that was not often assigned by humans (model 1) and generating a noun that was not selected by humans (model 2) on the inverse Jensen–Shannon divergence metric.}
  \label{fig:toy_example}
\end{figure*}

\subsection{QUANT Appendix}\label{appendix:quant}
\begin{figure*}[t]
\centering
  \includegraphics[width=0.9\linewidth]{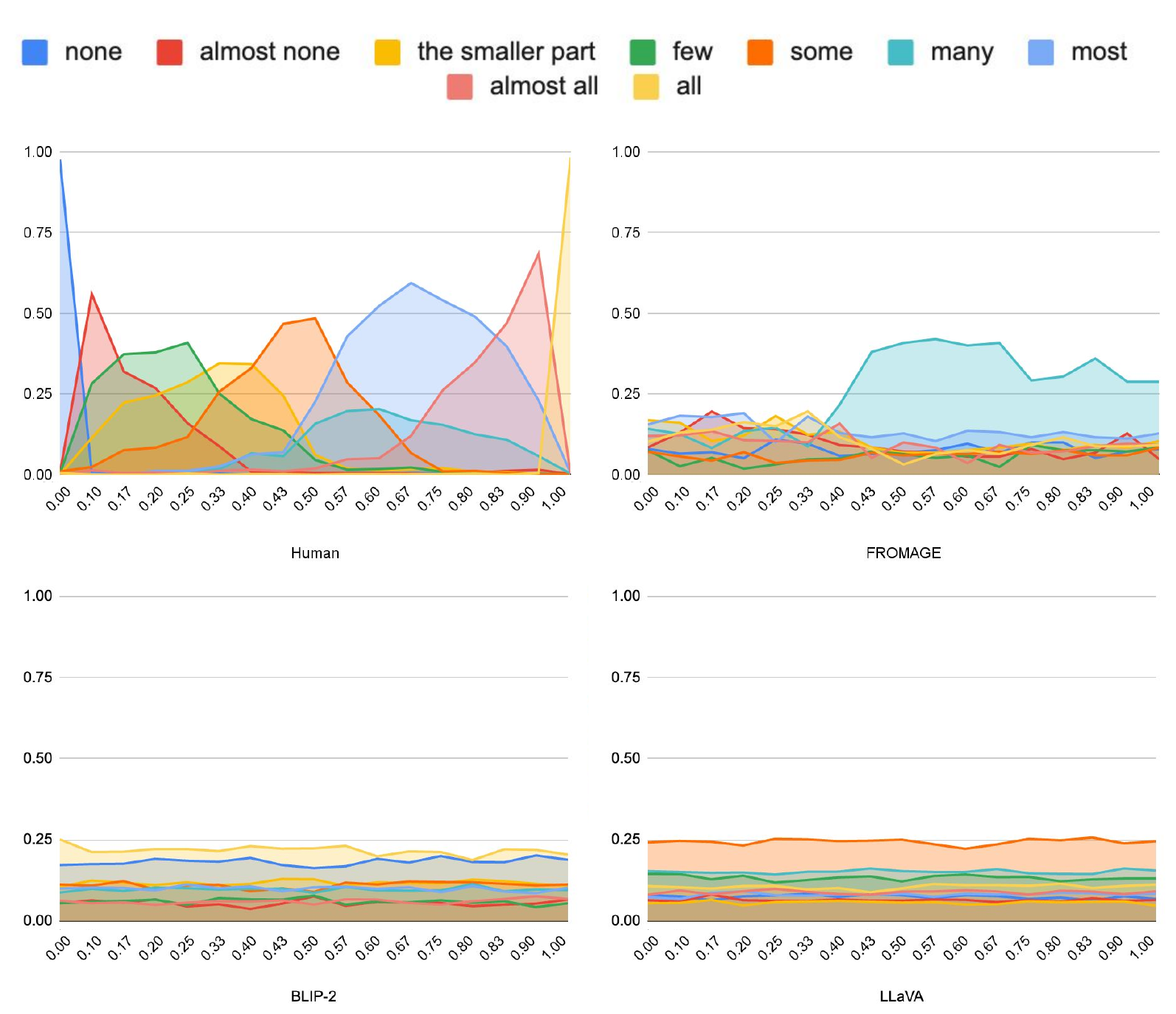}
  \caption{Density plot reporting the frequency distribution of responses for the 9 quantifiers (y-axis) against the proportion of targets in the scene (x-axis).}
  \label{fig:density_plots}
\end{figure*}

\begin{figure}[t]
\centering
  \includegraphics[width=1\linewidth]{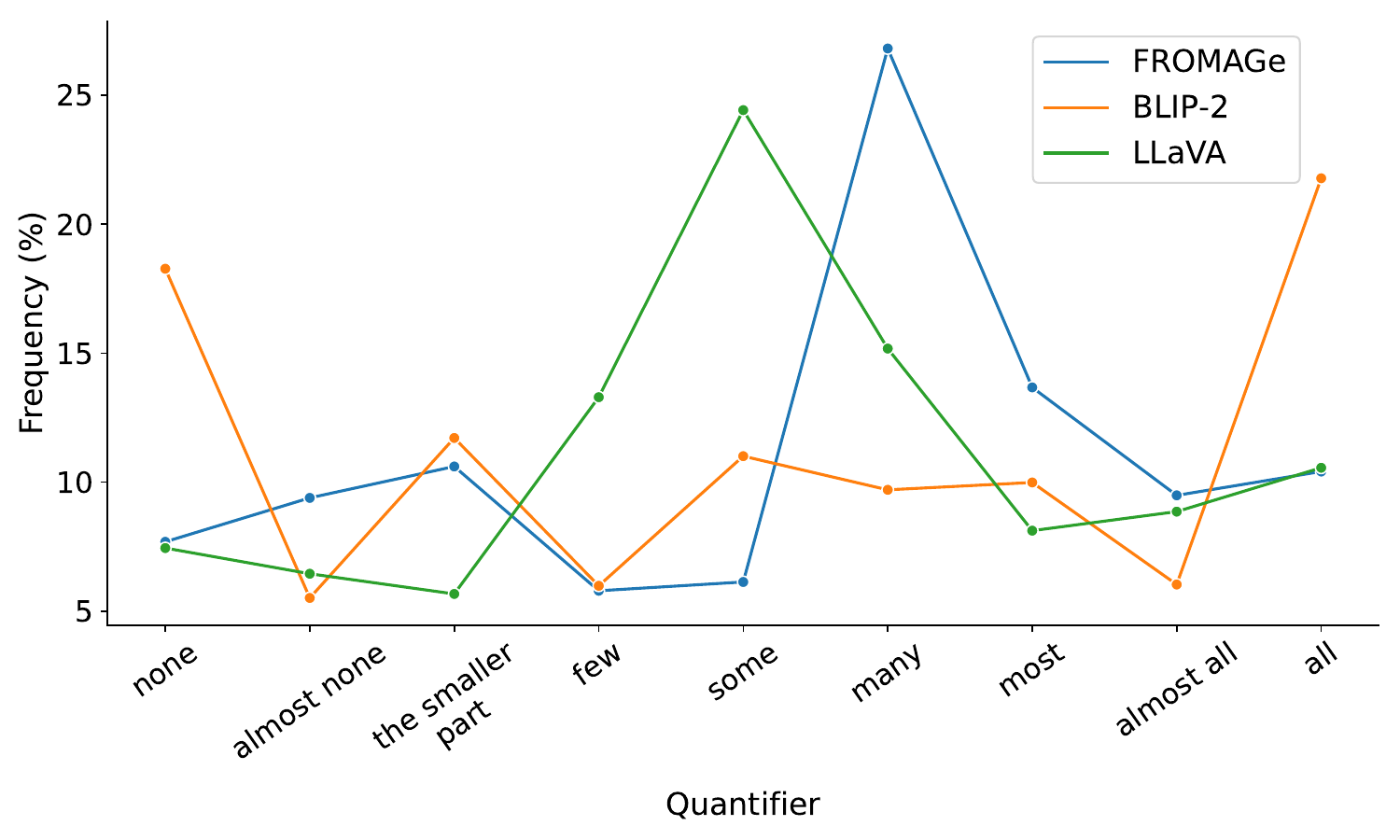}
  \caption{How often each quantifier (x-axis) is selected by the model (y-axis, expressed as \%), regardless of the proportion of targets (i.e., animals) in the image.}
  \label{fig:frequency_quantifiers}
\end{figure}

\begin{figure*}[t]
\centering
  \includegraphics[width=0.9\linewidth]{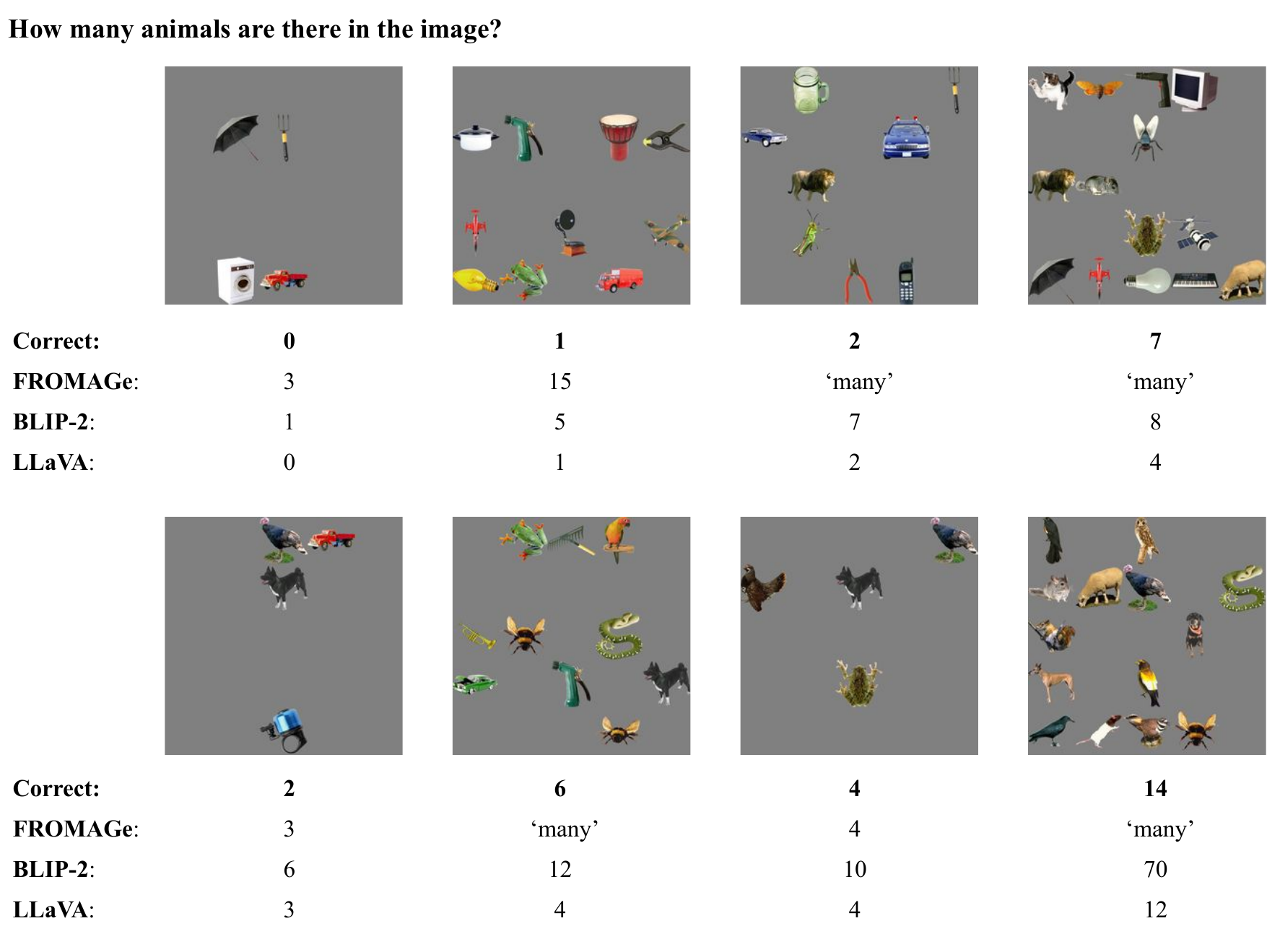}
  \caption{How many animals are there in the image? All models fail to successfully count the number of animals in the image. Note that models generally output a number but sometimes FROMAGe outputs the quantifier `many' which, interestingly, is the quantifier the model is strongly biased towards as illustrated in Figure \ref{fig:density_plots} and discussed in Section \ref{sec:curious_case_quantifiers}.}
  \label{fig:counting}
\end{figure*}

Figures \ref{fig:density_plots} and \ref{fig:frequency_quantifiers} show additional analyses on the QUANT dataset. They are discussed in Section \ref{sec:curious_case_quantifiers}. Figure \ref{fig:counting} shows some qualitative examples of the models' output when asked to answer the question `How many animals are there in the image?'. The images are randomly selected from QUANT. QUANT is released without a specific license. 

\begin{figure}[t]
\centering
  \includegraphics[width=1\linewidth]{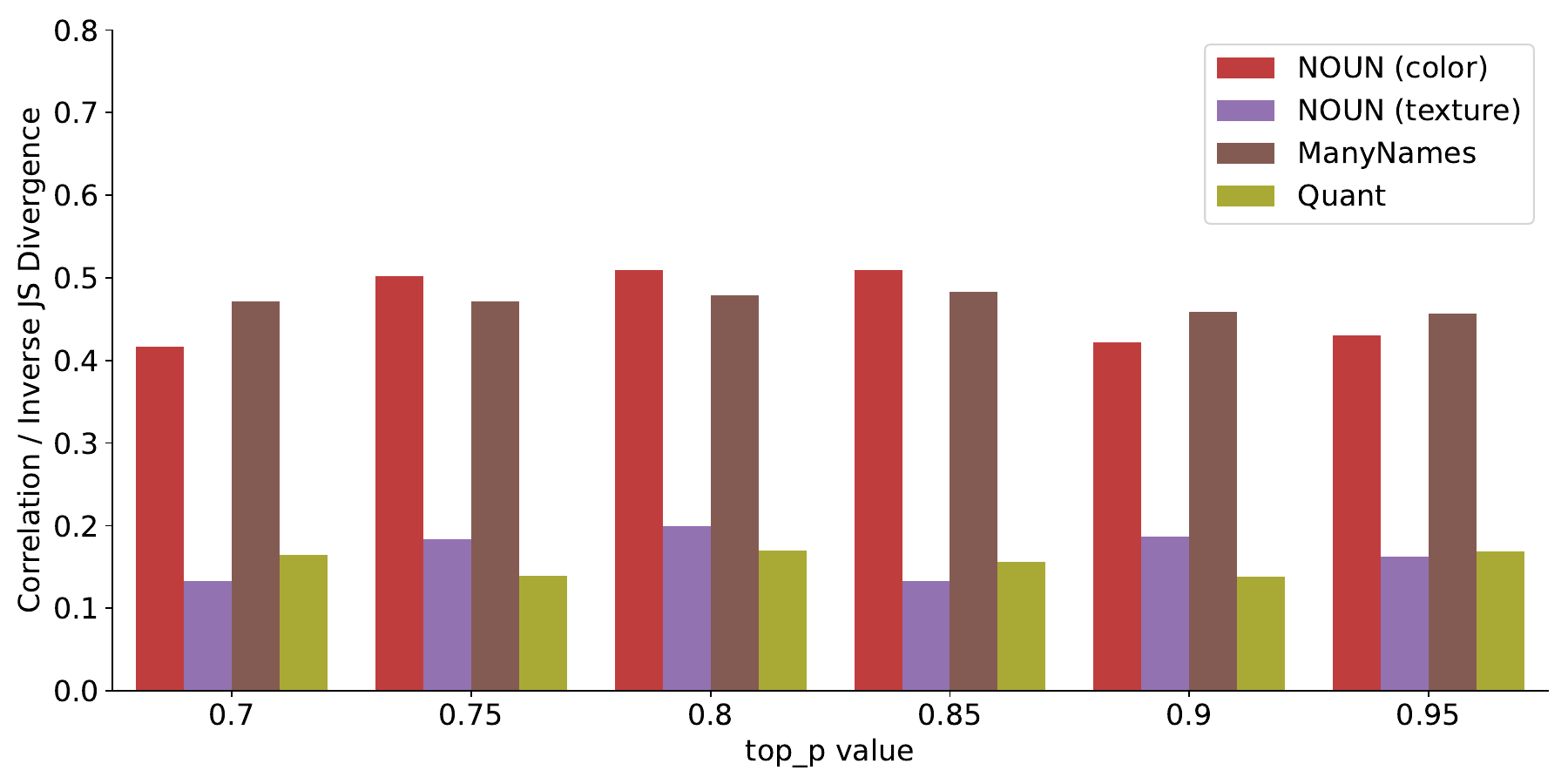}
  \caption{The role of different \textit{top\_p} values for nucleus sampling decoding using the LLaVa model.}
  \label{fig:top_p_values}
\end{figure}

\subsection{Models Appendix}\label{appendix:models}

While FROMAGe is trained with a contrastive learning objective for image captioning and it is shown to perform particularly well with longer textual contexts, BLIP-2 jointly optimizes three pre-training objectives that share the same input format and model parameters: image-text contrastive learning, image-grounded text generation, and image-text matching. 
%FROMAGe is shown to perform particularly well with longer textual contexts. 
The main innovation of LLaVA is the use GPT-4 generated visual instruction tuning data. Moreover, LLaVA has a simpler scheme to connect image and language representations compared to BLIP-2 and FROMAGe.  We used \texttt{blip2-opt-2.7b} and \texttt{llava-v1.5-7b}, while for FROMAGe we used the model made available by \citet{koh2023grounding}. FROMAGe and LLaVA are released with an Apache-2.0 license. BLIP-2 is distributed with BSD 3-Clause License. We run our experiments under the model license. 
%\alberto{Given space constraints, what should we highlight when describing/comparing the three models?}

\subsection{Generation Details}\label{appendix:generation}
\paragraph{The Effect of Different \textit{top\_p} Values} We experimented with various \textit{top\_p} values for nucleus sampling decoding. As illustrated in Figure \ref{fig:top_p_values} (showing the results for LLaVA, with similar results for the other models), we observe that this variable does not play a significant role in our experimental setup for all the tasks analyzed.

\paragraph{Different prompts} In our experiments, we prompted the models with the same instructions provided to human annotators during the collection of the different datasets analyzed. We also experimented with small variations of the above-mentioned prompts, such as `\textit{What is the object in the image?}' for the NOUN dataset, `\textit{Name the object in the red box with the most appropriate single name}' for ManyNames, and a more detailed instruction for QUANT, such as  `\textit{Carefully examine the image. Can you determine the proportion of animals present, compared to objects? Please select the most accurate answer from the options below}'. While we do not observe any significant difference between NOUN and ManyNames, the revised prompt for QUANT leads to a slight improvement in the model performance, with LLaVA reaching a correlation of 0.29. Still, the low absolute correlation coefficient highlights that computational models struggle to accurately assign quantifiers to visual scenes. This result demonstrates that the prompt may influence the performance of the models on this task. Although exploring which prompts work best was beyond the scope of this paper, we leave a systematic exploration of this aspect to future research.

% \paragraph{Model Output Examples} Figure \ref{fig:examples_appendix} shows some examples of the output generated by LLaVA (together with the instruction provided to the model). 
 
\subsection{Additional Details}
The data used in our work do not contain any information that names or uniquely identifies individual people or offensive content. FROMAGe has 5M trainable parameters and a total number of around 7.2B parameters. BLIP-2 has 188M trainable parameters and 2.7B total parameters. LLaVA has 7B parameters. All the models are evaluated on a single GPU (NVIDIA RTX A5000). We experimented with a few configurations of hyperparameters for nucleus sampling generation ( described in Section \ref{sec:experiments_results}). We did not find significant differences across different hyperparameters. We used the \texttt{SciPy} library (\url{https://scipy.org/}) to compute the correlation/divergence results. We used the NLTK library (\url{https://www.nltk.org/}) to extract nouns from the model output for ManyNames. 

\end{document}